%% file: arxiv.tex
\documentclass[runningheads]{llncs}
\usepackage[T1]{fontenc}
\usepackage{framed} 
\usepackage{hyperref}
\usepackage{longtable}
\usepackage{multirow}
\usepackage{graphicx}
\usepackage{subcaption}
\usepackage[table,xcdraw]{xcolor}
\usepackage[normalem]{ulem}
\usepackage{amsfonts} 
\usepackage{amsmath}
\usepackage{booktabs} 

\usepackage{makecell}
\usepackage{siunitx}
\usepackage{balance}
\usepackage[ruled,vlined]{algorithm2e}
\usepackage{colortbl}
\usepackage{xcolor}
\usepackage{amssymb}

\usepackage{algpseudocode}
\usepackage{wrapfig}

\newcommand{\methodname}{\textbf{Persistence}}
\newcommand{\alphareq}{$\alpha$-ReQ}
\newcommand{\coherence}{$\mu_0$-incoherence}
\newcommand{\rankme}{RankMe}
\newcommand{\stablerank}{StableRank}
\newcommand{\nesum}{NESum}
\newcommand{\selfcluster}{SelfCluster}
\newcommand{\pseudocond}{P-C number}

\begin{document}
\title{Topological Metric for Unsupervised Embedding Quality Evaluation}
\author{
	Aleksei Shestov\inst{1} \and
	Anton Klenitskiy\inst{1} \and
	Daria Denisova\inst{1} \and
	Amurkhan Dzagkoev\inst{1} \and
	Daniil Petrovich\inst{1} \and
	Andrey Savchenko\inst{1} \and
	Maksim Makarenko\inst{1}
}

\authorrunning{A. Shestov et al.}

\institute{
	Sber AI Laboratory, Moscow, Russia
}
\maketitle              

\vspace{-2.5em}
\begin{abstract}

\input{0_abstract}
\end{abstract}

\section{Introduction}
\label{sec:intro}
\input{1_intro}

\vspace{-0.6em}
\section{Proposed Approach}
\label{sec:approach}
\input{2_approach}

\section{Experiments}
\input{3_setup}
\input{4_results}
\vspace{-0.6em}
\section{Conclusion}
\label{sec:conclusion}
\input{5_conclusion}

\bibliographystyle{splncs04}
\bibliography{sample-base}

\end{document}

%% file: 0_abstract.tex
Modern representation learning increasingly relies on unsupervised and self-supervised methods trained on large-scale unlabeled data. While these approaches achieve impressive generalization across tasks and domains, evaluating embedding quality without labels remains an open challenge. In this work, we propose \methodname, a topology-aware metric based on persistent homology that quantifies the geometric structure and topological richness of embedding spaces in a fully unsupervised manner. Unlike metrics that assume linear separability or rely on covariance structure, \methodname\ captures global and multi-scale organization.
Empirical results across diverse domains show that \methodname\ consistently achieves top-tier correlations with downstream performance, outperforming existing unsupervised metrics and enabling reliable model and hyperparameter selection.

%% file: 1_intro.tex
Modern representation learning increasingly relies on unsupervised and self-supervised methods that leverage large-scale unlabeled data through proxy or contrastive objectives~\cite{chen2020simple_icml,he2020moco_cvpr,grill2020byol_neurips,he2022mae_cvpr,jing2019self,gui2023survey}. In parallel, zero-shot models have shown strong generalization to unseen domains and tasks~\cite{brown2020gpt3_neurips,jia2021align_icml,radford2021clip_icml}. As these approaches operate without labels, a key challenge is to assess embedding quality and guide model selection without supervised validation~\cite{agrawal2022alpha,tsitsulin2023unsup_evaluation_icmlw}. Reliable label-free metrics are thus essential for scalable and autonomous learning, including applications in recommender systems and information retrieval.

Embeddings must preserve the geometric structure of the data manifold to perform well on unseen tasks. Intuitively, collapsing distinct regions or losing multi-scale relationships limits downstream performance. Conversely, embeddings retaining intrinsic geometric complexity better support diverse tasks. This motivates evaluating embeddings by the richness of their geometric structure.

Measuring geometric structure in a principled, model-agnostic way is challenging. Existing metrics rely on simplifying assumptions: \rankme~\cite{garrido2023rankme} assumes linear subspaces, \alphareq~\cite{agrawal2022alpha} and \nesum~\cite{he2022exploring} use covariance models that fail on multimodal data, and \coherence~\cite{tsitsulin2023unsupervised} presumes linear separability. \selfcluster~\cite{tsitsulin2023unsupervised} suits contrastive setups but poorly generalizes beyond them. This highlights the need for more holistic, model-agnostic evaluation.

Topological methods naturally address these limitations by capturing global and multi-scale data structure through connected components, loops, and cavities that persist across scales~\cite{carlsson_topology_2009,edelsbrunner2010computational}. Intuitively, embeddings that preserve rich data structure exhibit long-lived topological features. Persistent homology, the core tool of Topological Data Analysis (TDA), quantifies such features via persistence diagrams~\cite{ghrist_barcodes_2008,otter2017roadmap} and has been successfully applied in biology~\cite{nicolau2011topology}, materials science~\cite{hiraoka2016hierarchical}, and neuroscience~\cite{sizemore2018clique}. However, its use for evaluating unsupervised embeddings remains largely unexplored.

To address this gap, we introduce \methodname\footnote{CODE: \url{https://anonymous.4open.science/r/topo_metrics-94D6/}}, a topological metric for unsupervised embedding quality evaluation. \methodname\ constructs a Vietoris–Rips complex over the embedding point cloud and computes persistence diagrams for homology groups \(H_0\) and \(H_1\). The total persistence, defined as the sum of topological feature lifetimes, quantifies the geometric richness of the embedding space (see Section~\ref{sec:approach}). Unlike prior methods, \methodname\ makes no assumptions about embedding geometry, downstream tasks, or objectives, enabling a model-agnostic evaluation across domains. In extensive benchmarks covering financial analytics, user behavioral modeling (UBM), and collaborative filtering, \methodname\ achieves the highest or near-highest correlations across all configurations, with mean Spearman correlations of up to 0.84 in UBM and 0.76 in collaborative filtering. In the model selection per epoch setting, it reaches Spearman 0.61 and quality 0.71, outperforming competing metrics such as \rankme, \alphareq, and \coherence. Overall, \methodname\ consistently aligns with supervised performance, generalizes across different domains and tasks, and establishes a new state of the art in geometry-aware, label-free embedding evaluation.


%% file: 2_approach.tex

\vspace{-0.6em}
\begin{figure}[htbp]
  \centering
  \includegraphics[width=\linewidth]{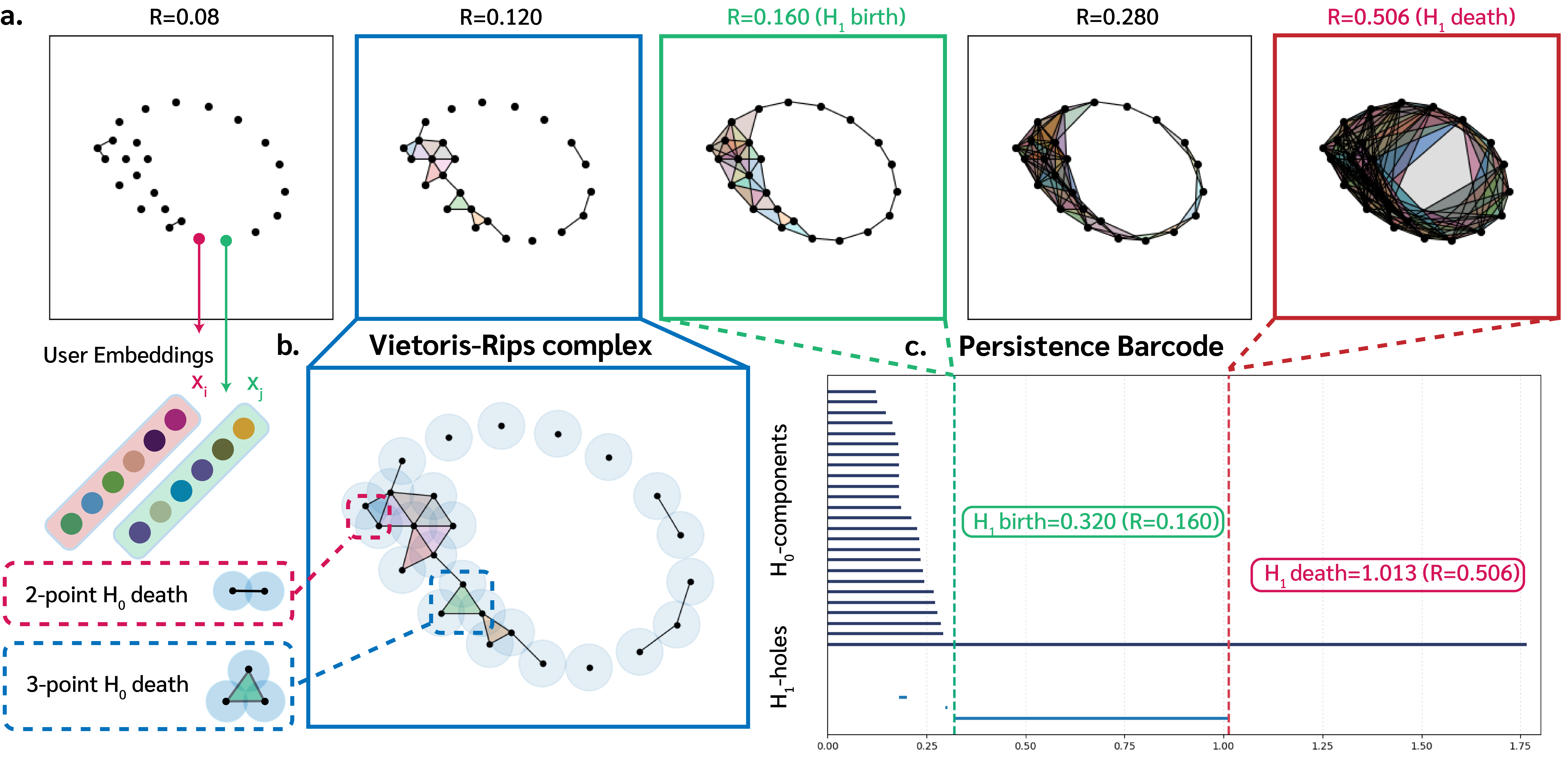}
  \caption{
  \textbf{Vietoris–Rips filtration and persistence barcode.}
  \textbf{(a)}~Evolution of the complex as the radius~$r$ grows, showing transitions from isolated points to a connected, contractible structure. 
  \textbf{(b)}~Complex at~$r=0.12$ with metric spheres, illustrating \emph{two-point} and \emph{three-point} $H_0$ deaths. 
  \textbf{(c)}~Corresponding persistence barcode with $H_0$ and $H_1$ intervals across scales.
  }
  \label{fig:vietoris_rips_barcode}
\end{figure}

\noindent\textbf{Topology-Aware Metric for Embedding Spaces.} We introduce a topology-aware metric based on persistent homology~\cite{ghrist_barcodes_2008} to quantify the intrinsic structure of embedding spaces. Figure~\ref{fig:vietoris_rips_barcode} illustrates the computation of persistent homology for a set of embedding vectors.
Given a point cloud of embeddings
$X_n = \{x_1, x_2, \ldots, x_n\} \subset \mathbb{R}^d$, 
we construct a Vietoris–Rips filtration to track how topological features emerge and vanish as the connectivity radius~$r$ increases.
As $r$ grows, isolated points begin to connect via edges, forming larger connected components and transient one-dimensional loops (Fig.~\ref{fig:vietoris_rips_barcode}a).
At a specific scale, for instance $r = 0.12$, we visualize the resulting simplicial complex together with metric spheres centered at the embedding points.
When two spheres first intersect, their corresponding points are joined by an edge, representing a two-point death of an $H_0$ feature (a component merge).
When three spheres intersect simultaneously, the corresponding points form a filled triangle (a 2-simplex), leading to a three-point death, which fills a loop and terminates an $H_1$ feature (Fig.~\ref{fig:vietoris_rips_barcode}b).
The persistence barcode summarizes these topological transitions across scales: each horizontal bar corresponds to a topological feature that appears at its birth value~$b_i$ and disappears at its death value~$d_i$ - either by merging with another component ($H_0$) or being filled by higher-dimensional simplices ($H_1$) (Fig.~\ref{fig:vietoris_rips_barcode}c).
Finally, the total persistence in homology dimension~$k$ quantifies the overall topological complexity as the normalized sum of feature lifetimes:
\begin{equation}
    \text{Persistence}_k(X_n) = 
    \frac{\sum_{(b_i, d_i) \in H_k} (d_i - b_i)}
         {\max_{i,j} \|x_i - x_j\|},
\end{equation}
where $H_k$ denotes the set of persistence pairs $(b_i, d_i)$ in dimension~$k$, and the denominator normalizes by the maximum pairwise distance between embeddings in $X_n$.  

\noindent\textbf{Linking Persistence to Intrinsic Dimension.} 
The core intuition behind the proposed approach is that the geometry of an embedding determines how much of the original data manifold it preserves. 
When embeddings maintain rich topological structure, connected components and loops persist over larger connectivity scales, leading to higher total persistence. 
This persistence serves as a proxy for the intrinsic dimension of the embedding manifold. 
Let $X_n = \{x_1, x_2, \ldots, x_n\} \subset \mathbb{R}^d$ denote a sample of $n$ points drawn uniformly from a smooth $d$-dimensional manifold $M$ with volume $\mathrm{Vol}(M)$ and diameter $\mathrm{Diam}(M)$. 
The expected zero-dimensional persistence then scales as
\begin{equation}
\mathbb{E}[\mathrm{Persistence}_0(X_n)] 
= \alpha(d)\, n^{1 - \frac{1}{d}} 
\frac{\mathrm{Vol}(M)^{1/d}}{\mathrm{Diam}(M)} 
+ o\!\left(n^{1 - \frac{1}{d}}\right),    
\end{equation}

where $\alpha(d)$ is a dimension-dependent constant and $o(\cdot)$ denotes a higher-order remainder term~\cite{yukich2006probability}. 
Higher intrinsic dimension $d$ results in more gradual topological merging, reflected in larger persistence values. 
Conversely, embeddings with dimension lower than that of the data manifold collapse regions of $M$, reducing the number of distinguishable events and degrading classification accuracy. 
Thus, high total persistence indicates that the embedding retains sufficient intrinsic dimensionality to preserve discriminative structure, explaining its observed correlation with improved downstream performance.

%% file: 3_setup.tex
\subsection{Experimental Setup}
\textbf{Domains and Datasets}. We evaluate \methodname\ across three application families that jointly span structured behavioral data, event-sequence modeling, and user–item interactions: financial analytics, user behavioral modeling (UBM), and recommendation systems. This setup ensures a diverse evaluation covering both domain-specific prediction tasks and general-purpose interaction modeling.
In the financial analytics family, we use two real-world banking datasets for gender and age group prediction~\cite{dllllb2024gender,dllllb2024agegroup}. The UBM family is represented by the Synerise dataset~\cite{dkabrowski2025synerise} from the RecSys Challenge 2025~\cite{dabrowski2025recsys}, which includes rich multi-event user logs such as purchases, cart additions, and page visits. The goal is to create universal user representations applicable across different tasks including churn, product popensity, and category propensity prediction. Finally, the recommendation family leverages the MovieLens-20M dataset~\cite{harper2015movielens} to evaluate performance on user–item collaborative filtering.

\noindent \textbf{Embedding Training Methods}.
In financial analytics, user embeddings are learned via the self-supervised CoLES framework~\cite{babaev2022coles}, which processes temporal transaction sequences through GRU/LSTM encoders trained with a margin-based contrastive loss~\cite{hadsell2006dimensionality}. For user behavioral modeling, embeddings are trained by a GRU-based event sequence autoencoder~\cite{klenitskiy2025encode}. Collaborative filtering embeddings are derived from matrix factorization methods including implicit ALS (iALS)~\cite{hu2008collaborative} and Bayesian Personalized Ranking (BPR)~\cite{rendle2012bpr}.

\noindent \textbf{Baseline Metrics and Evaluation}.
We compare the topology-based \methodname\ metric against spectral metrics (\rankme~\cite{garrido2023rankme}, \alphareq~\cite{agrawal2022alpha}, \nesum~\cite{he2022exploring}, \coherence, \stablerank, \pseudocond~\cite{tsitsulin2023unsupervised}) and clustering-based \selfcluster~\cite{tsitsulin2023unsupervised}. Downstream performance is measured using ROC AUC and accuracy for gender and age prediction, ROC AUC in combination with novelty and diversity for RecSys Challenge tasks~\cite{dabrowski2025recsys}, and NDCG@10 for recommendation tasks.
To assess how well unsupervised metrics reflect downstream quality, we compute Pearson and Spearman correlations between unsupervised and downstream metrics, and evaluate downstream quality scores achieved by selecting models solely based on unsupervised metric values.

%% file: 4_results.tex
\vspace{-0.6em}
\subsection{Results}
\textbf{Hyperparameter Optimization.} In this section we benchmark unsupervised metrics within the hyperparameter search task, considering three configurations covering financial analytics, user behavioral modeling, and collaborative filtering.

\input{grid_search_all_models}

 For the \textbf{financial analytics} configuration, we vary batch size (16-256), learning rate ($10^{-5}$-0.05), temporal splits (3-7), transaction limits, embedding size (32-1024), hidden size (64-4096), loss function (Contrastive, Barlow Twins, VICReg), and encoder type (GRU, LSTM), training each configuration for 30 epochs. The correlations and downstream quality are averaged across two datasets and two downstream classifiers (CatBoost and MLP). \methodname\ achieves top-1/2 results with a Pearson correlation of 0.671, Spearman correlation of 0.560, and a quality score of 0.728 (Table~\ref{table:grid_search_all_models}). The slightly higher Pearson value of 0.710 achieved by \nesum\ is accompanied by lower Spearman (0.529) and quality (0.697), indicating weaker overall consistency between unsupervised and downstream performance.

For the \textbf{user behavioral modeling} (UBM) configuration, a GRU-based autoencoder is used with variations in embedding size (64-512), dropout (0.1, 0.3), number of layers (1-4), and sequence length (64-256). The correlations are computed for total sum of scores across all downstream tasks, and this sum is reported as downstream quality.
In this evaluation, \methodname\ provides the top-1 performance across all metrics, achieving a Pearson correlation of 0.861, Spearman correlation of 0.840, and a quality score of 2.263 (Table~\ref{table:grid_search_all_models}). \nesum\ attains comparable quality (2.263) but substantially lower correlations (Pearson 0.638, Spearman 0.303).

\input{recsys_all_corr}

For the \textbf{collaborative filtering} setup, the downstream objective is explicitly defined during training, and embeddings are optimized for a single known target rather than for diverse unseen tasks. In this setup, we use absolute values of correlations since lower unsupervised metric values correspond to better model performance. For iALS, we vary latent factors (16-512), regularization (0.01-10), and confidence scaling $\alpha$ (0.03-1); for BPR, we vary latent factors (16 to 512), regularization (0.001-0.1), and learning rate (0.0003-0.03). Table~\ref{tab:recsys_result} reports mean Pearson and Spearman correlations between unsupervised metrics and downstream NDCG@10 performance, computed separately for user and item embeddings, along with their average. \methodname\ achieves the highest overall agreement, with average Pearson and Spearman correlations of 0.778 and 0.763, respectively, and item-level values of 0.893 and 0.792. \nesum\ provides the next best rank correlation (average Spearman 0.713), while \stablerank\ shows a strong user Pearson correlation of 0.744 but lower item consistency. 

\input{epoch_choose}

\noindent\textbf{Optimal epoch selection.} 
In this section we benchmark unsupervised metrics for optimal epoch selection in the financial analytics configuration. Model checkpoints are evaluated across training epochs with fixed hyperparameters, and performance indicators are averaged over all hyperparameter configurations.
 Table~\ref{table:epoch_choose} summarizes mean Pearson and Spearman correlations, along with the downstream quality metric, averaged across the Gender and Age datasets and two downstream classifiers, CatBoost and MLP. \methodname\ achieves the highest correlations and quality with Pearson 0.691, Spearman 0.609, and quality 0.713. \nesum\ attains comparable quality (0.713) but lower correlations (Pearson 0.665, Spearman 0.532), while \rankme\ shows similar Spearman (0.532) but lower quality (0.709).

%% file: grid_search_all_models.tex
\begin{table}[!t]
\small
\centering
\caption{Financial analytics and UBM. Mean Pearson, Spearman, and quality scores for configurations selected by each unsupervised metric. Top-3 absolute values per column are color-coded: \textcolor{blue}{blue} (1st), \textcolor{green!50!black}{green} (2nd), \textcolor{orange!70!black}{orange} (3rd).}
\label{table:grid_search_all_models}
\begin{tabular}{lccc|ccc}
\toprule
\textbf{Configuration} & \multicolumn{3}{c}{\textbf{Financial analytics}} & \multicolumn{3}{c}{\textbf{Behavioral modeling}} \\
\cmidrule(lr){2-4} \cmidrule(lr){5-7}
\textbf{Metric} & Pearson & Spearman & Quality & Pearson & Spearman & Quality \\
\midrule
\methodname   & \cellcolor{green!15}0.671 & \cellcolor{blue!15}0.560 & \cellcolor{blue!15}0.728 & \cellcolor{blue!15}0.861 & \cellcolor{blue!15}0.840 & \cellcolor{blue!15}2.263 \\
\nesum        & \cellcolor{blue!15}0.710  & \cellcolor{green!15}0.529 & 0.697                     & 0.638                     & 0.303                     & \cellcolor{blue!15}2.263 \\
\coherence    & \cellcolor{orange!15}0.409& 0.324                     & 0.653                     & \cellcolor{green!15}{0.818} & \cellcolor{orange!15}{0.479} & \cellcolor{orange!15}2.214 \\
\selfcluster  & 0.338                      & 0.358                     & 0.626                     & \cellcolor{orange!15}{0.646} & 0.345                    & 2.192 \\
\rankme       & 0.274                      & \cellcolor{orange!15}0.525& \cellcolor{green!15}0.710 & 0.508                     & \cellcolor{green!15}0.693 & \cellcolor{green!15}2.241 \\
\stablerank   & 0.130                      & 0.367                     & 0.635                     & 0.342                     & 0.345                     & 2.192 \\
\pseudocond   & 0.120                      & 0.263                     & \cellcolor{orange!15}0.700& 0.163                     & 0.030                     & 2.197 \\
\alphareq     & 0.160                     & 0.273                     & 0.644                     & 0.264                    & 0.040                    & \cellcolor{green!15}2.241 \\
\bottomrule
\end{tabular}
\vspace{-0.6em}
\end{table}

%% file: recsys_all_corr.tex
\begin{table*}[!ht]
\vspace{-1em}
\small
\centering
\caption{Collaborative filtering results. Mean correlations for unsupervised metrics on user and item embeddings, and their average.}

\label{tab:recsys_result}
\begin{tabular}{lcc|cc|cc}
\toprule
\textbf{Type} & \multicolumn{2}{c}{\textbf{User}} & \multicolumn{2}{c}{\textbf{Item}} & \multicolumn{2}{c}{\textbf{Average}} \\
\cmidrule(lr){2-3}\cmidrule(lr){4-5}\cmidrule(lr){6-7}
\textbf{Correlation} & Pearson & Spearman & Pearson & Spearman & Pearson & Spearman \\
\midrule
\methodname   & \cellcolor{orange!15}0.617 & \cellcolor{green!15}0.678 & \cellcolor{blue!15}0.893 & \cellcolor{blue!15}0.792 & \cellcolor{blue!15}0.778 & \cellcolor{blue!15}0.763 \\
\nesum        & 0.434 & \cellcolor{blue!15}0.705 & \cellcolor{green!15}0.699 & \cellcolor{orange!15}0.565 & 0.437 & \cellcolor{green!15}0.713 \\
\rankme       & \cellcolor{green!15}0.676 & 0.588 & \cellcolor{orange!15}0.651 & \cellcolor{green!15}0.647 & \cellcolor{orange!15} 0.669 &\cellcolor{orange!15}0.660 \\
\stablerank   & \cellcolor{blue!15}0.744 & 0.583 & 0.615 & 0.401 & \cellcolor{green!15}0.697 & 0.494 \\
\alphareq     & 0.299 & 0.490 & 0.061 & 0.309 & 0.199 & 0.447 \\
\pseudocond   & 0.414 & 0.232 & 0.496 & 0.431 & 0.528 & 0.425 \\
\coherence    & 0.292 & 0.406 & 0.191 & 0.257 & 0.220 & 0.309 \\
\selfcluster  & 0.512 & \cellcolor{orange!15}0.605 & 0.393 & 0.234 & 0.300 & 0.280 \\
\bottomrule
\end{tabular}
\vspace{-0.6em}
\end{table*}

%% file: epoch_choose.tex
\begin{wraptable}{r}{7.5cm}
\vspace{-0.6em}
\small
\centering
\caption{Epoch selection: correlations and quality averaged over Gender/Age, CatBoost/MLP. }
\label{table:epoch_choose}
\begin{tabular}{lccc}
\toprule
\textbf{Metric} & \textbf{Pearson} & \textbf{Spearman} & \textbf{Quality} \\
\midrule
\methodname   & \cellcolor{blue!15}0.691 & \cellcolor{blue!15}0.609 & \cellcolor{blue!15}0.713 \\
\nesum        & \cellcolor{green!15}0.665 & \cellcolor{green!15}0.532 & \cellcolor{blue!15}0.713 \\
\rankme       & \cellcolor{orange!15}0.613 & \cellcolor{green!15}0.532 & \cellcolor{green!15}0.709 \\
\selfcluster  & 0.597 & \cellcolor{orange!15}0.463 & \cellcolor{orange!15}0.707 \\
\stablerank   & 0.595 & \cellcolor{orange!15}0.463 & 0.708 \\
\coherence    & 0.454 & 0.304 & 0.708 \\
\pseudocond   & 0.247 & 0.226 & 0.692 \\
\alphareq     & 0.115 & 0.229 & 0.681 \\
\bottomrule
\end{tabular}
\vspace{-0.6em}
\end{wraptable}

%% file: 5_conclusion.tex
We presented \methodname, a topology-aware metric for unsupervised evaluation of embedding quality based on persistent homology. By constructing Vietoris–Rips complexes and quantifying the total persistence of topological features, \methodname\ provides a geometry-sensitive measure of representation richness that requires no labels or task-specific assumptions. 

Extensive experiments across three configurations -- financial analytics, user behavioral modeling (UBM), and collaborative filtering -- demonstrate that \methodname\ consistently ranks among the top-1 or top-2 unsupervised metrics in correlation with downstream performance. It reliably supports both hyperparameter search and model selection per epoch, outperforming existing approaches such as \rankme, \alphareq, and \coherence. Overall, the proposed approach demonstrates that persistent homology provides a practical foundation for label-free evaluation of learned representations across diverse data modalities.

Future research directions may include deepening the theoretical link between persistent homology and embedding generalization, and extending the proposed framework to vision, language, and multimodal domains with higher geometric complexity.


%% file: sample-base.bib
@inproceedings{hu2008collaborative,
  title        = {Collaborative Filtering for Implicit Feedback Datasets},
  author       = {Hu, Yifan and Koren, Yehuda and Volinsky, Chris},
  booktitle    = {Proceedings of the 2008 Eighth IEEE International Conference on Data Mining (ICDM)},
  year         = {2008},
  pages        = {263--272},
  publisher    = {IEEE},
  doi          = {10.1109/ICDM.2008.22},
  url          = {https://doi.org/10.1109/ICDM.2008.22}
}

@book{edelsbrunner2010computational,
  title     = {Computational Topology: An Introduction},
  author    = {Edelsbrunner, Herbert and Harer, John L.},
  year      = {2010},
  publisher = {American Mathematical Society},
  address   = {Providence, RI},
  isbn      = {978-0-8218-4925-5}
}

@article{otter2017roadmap,
  title   = {A Roadmap for the Computation of Persistent Homology},
  author  = {Otter, Nina and Porter, Mason A. and Tillmann, Ulrike and Grindrod, Peter and Harrington, Heather A.},
  journal = {EPJ Data Science},
  volume  = {6},
  number  = {1},
  pages   = {17},
  year    = {2017},
  doi     = {10.1140/epjds/s13688-017-0109-5},
  url     = {https://epjdatascience.springeropen.com/articles/10.1140/epjds/s13688-017-0109-5}
}

@article{nicolau2011topology,
  title   = {Topology-Based Data Analysis Identifies a Subgroup of Breast Cancers with a Unique Mutational Profile and Excellent Survival},
  author  = {Nicolau, Monica and Levine, Arnold J. and Carlsson, Gunnar},
  journal = {Proceedings of the National Academy of Sciences},
  volume  = {108},
  number  = {17},
  pages   = {7265--7270},
  year    = {2011},
  doi     = {10.1073/pnas.1102826108},
  url     = {https://doi.org/10.1073/pnas.1102826108}
}

@article{hiraoka2016hierarchical,
  title   = {Hierarchical Structures of Amorphous Solids Characterized by Persistent Homology},
  author  = {Hiraoka, Yasuaki and Nakamura, Tomoyuki and Hirata, Akihiko and Escolar, Ezra G. and Matsue, Kazushi and Nishiura, Yasumasa},
  journal = {Proceedings of the National Academy of Sciences},
  volume  = {113},
  number  = {26},
  pages   = {7035--7040},
  year    = {2016},
  doi     = {10.1073/pnas.1520877113},
  url     = {https://doi.org/10.1073/pnas.1520877113}
}

@article{sizemore2018clique,
  title   = {Cliques and Cavities in the Human Connectome},
  author  = {Sizemore, Ann E. and Giusti, Chad and Kahn, Ariel and Vettel, Jean M. and Betzel, Richard F. and Bassett, Danielle S.},
  journal = {Journal of Computational Neuroscience},
  volume  = {44},
  number  = {1},
  pages   = {115--145},
  year    = {2018},
  doi     = {10.1007/s10827-017-0672-6},
  url     = {https://doi.org/10.1007/s10827-017-0672-6}
}

@inproceedings{chen2020simple_icml,
  title     = {A Simple Framework for Contrastive Learning of Visual Representations},
  author    = {Chen, Ting and Kornblith, Simon and Norouzi, Mohammad and Hinton, Geoffrey},
  booktitle = {Proceedings of the 37th International Conference on Machine Learning (ICML)},
  year      = {2020},
  pages     = {1597--1607},
  publisher = {PMLR},
  url       = {https://proceedings.mlr.press/v119/chen20j.html}
}

@inproceedings{he2020moco_cvpr,
  title     = {Momentum Contrast for Unsupervised Visual Representation Learning},
  author    = {He, Kaiming and Fan, Haoqi and Wu, Yuxin and Xie, Saining and Girshick, Ross},
  booktitle = {Proceedings of the IEEE/CVF Conference on Computer Vision and Pattern Recognition (CVPR)},
  year      = {2020},
  pages     = {9729--9738},
  doi       = {10.1109/CVPR42600.2020.00974}
}

@inproceedings{grill2020byol_neurips,
  title     = {Bootstrap Your Own Latent: A New Approach to Self-Supervised Learning},
  author    = {Grill, Jean-Bastien and Strub, Florian and Altch{\'e}, Florent and Tallec, Corentin and Richemond, Pierre H. and Buchatskaya, Elena and Doersch, Carl and Pires, Bernardo A. and Guo, Zhaohan Daniel and Azar, Mohammad Gheshlaghi and Piot, Bilal and Kavukcuoglu, Koray and Munos, Remi and Valko, Michal},
  booktitle = {Advances in Neural Information Processing Systems (NeurIPS)},
  year      = {2020},
  volume    = {33},
  pages     = {21271--21284},
  url       = {https://papers.nips.cc/paper/2020/file/f3ada80d5c4ee70142b17b8192b2958e-Paper.pdf}
}

@inproceedings{he2022mae_cvpr,
  title     = {Masked Autoencoders Are Scalable Vision Learners},
  author    = {He, Kaiming and Chen, Xinlei and Xie, Saining and Li, Yanghao and Doll{\'a}r, Piotr and Girshick, Ross},
  booktitle = {Proceedings of the IEEE/CVF Conference on Computer Vision and Pattern Recognition (CVPR)},
  year      = {2022},
  pages     = {16000--16009},
  doi       = {10.1109/CVPR52688.2022.01548}
}

@inproceedings{radford2021clip_icml,
  title     = {Learning Transferable Visual Models From Natural Language Supervision},
  author    = {Radford, Alec and Kim, Jong Wook and Hallacy, Chris and Ramesh, Aditya and Goh, Gabriel and Agarwal, Sandhini and Sastry, Girish and Askell, Amanda and Mishkin, Pamela and Clark, Jack and Krueger, Gretchen and Sutskever, Ilya},
  booktitle = {Proceedings of the 38th International Conference on Machine Learning (ICML)},
  year      = {2021},
  pages     = {8748--8763},
  publisher = {PMLR},
  url       = {https://proceedings.mlr.press/v139/radford21a.html}
}

@inproceedings{jia2021align_icml,
  title     = {Scaling Up Visual and Vision-Language Representation Learning With Noisy Text Supervision},
  author    = {Jia, Chao and Yang, Yinfei and Xia, Ye and Chen, Yi-Ting and Parekh, Zarana and Pham, Hieu and Le, Quoc V. and Sung, Yunhsuan and Li, Zhen and Duerig, Tom},
  booktitle = {Proceedings of the 38th International Conference on Machine Learning (ICML)},
  year      = {2021},
  pages     = {4904--4916},
  publisher = {PMLR},
  url       = {https://proceedings.mlr.press/v139/jia21b.html}
}

@article{brown2020gpt3_neurips,
  title     = {Language Models are Few-Shot Learners},
  author    = {Brown, Tom B. and Mann, Benjamin and Ryder, Nick and Subbiah, Melanie and Kaplan, Jared D. and Dhariwal, Prafulla and Neelakantan, Arvind and Shyam, Pranav and Sastry, Girish and Askell, Amanda and Agarwal, Sandhini and Herbert-Voss, Ariel and Krueger, Gretchen and Henighan, Tom and Child, Rewon and Ramesh, Aditya and Ziegler, Daniel M. and Wu, Jeffrey and Winter, Clemens and Hesse, Chris and Chen, Mark and Sigler, Eric and Litwin, Mateusz and Gray, Scott and Chess, Benjamin and Clark, Jack and Berner, Christopher and McCandlish, Sam and Radford, Alec and Sutskever, Ilya and Amodei, Dario},
  journal   = {Advances in Neural Information Processing Systems (NeurIPS)},
  volume    = {33},
  pages     = {1877--1901},
  year      = {2020},
  url       = {https://papers.nips.cc/paper/2020/file/1457c0d6bfcb4967418bfb8ac142f64a-Paper.pdf}
}

@article{tsitsulin2023unsup_evaluation_icmlw,
  title   = {Unsupervised Evaluation of Representation Learning},
  author  = {Tsitsulin, Anton and Palowitch, John and Perozzi, Bryan and Beygelzimer, Alina},
  journal = {ICML Workshop on Unsupervised Learning},
  year    = {2023},
  url     = {https://openreview.net/forum?id=bVp2xljE5d}
}

@article{jing2019self,
  title   = {Self-supervised visual feature learning with deep neural networks: A survey},
  author  = {Jing, Longlong and Tian, Yingli},
  journal = {IEEE Transactions on Pattern Analysis and Machine Intelligence},
  year    = {2020},
  volume  = {43},
  number  = {11},
  pages   = {4037--4058},
  doi     = {10.1109/TPAMI.2020.2992393}
}

@article{gui2023survey,
  title   = {A Survey on Self-Supervised Learning: Algorithms, Applications, and Future Trends},
  author  = {Gui, Jie and Chen, Tuo and Zhang, Jing and Cao, Qiong and Sun, Zhenan and Luo, Hao and Tao, Dacheng},
  journal = {IEEE Transactions on Pattern Analysis and Machine Intelligence},
  year    = {2023},
  doi     = {10.1109/TPAMI.2023.3239877}
}

@inproceedings{garrido2023rankme,
    author    = "Garrido, Q. and Balestriero, R. and Najman, L. and LeCun, Y.",
    title     = "RankMe: Assessing the downstream performance of pretrained self-supervised representations by their rank",
    booktitle = "Proceedings of the 40th International Conference on Machine Learning",
    pages     = "10929--10974",
    year      = "2023",
    publisher = "PMLR",
    url       = "https://proceedings.mlr.press/v202/garrido23a/garrido23a.pdf",
}

@inproceedings{he2022exploring,
  title={Exploring the gap between collapsed \& whitened features in self-supervised learning},
  author={He, Bobby and Ozay, Mete},
  booktitle={International Conference on Machine Learning},
  pages={8613--8634},
  year={2022},
  organization={PMLR}
}

@article{agrawal2022alpha,
  title={$\alpha $-ReQ: Assessing Representation Quality in Self-Supervised Learning by measuring eigenspectrum decay},
  author={Agrawal, Kumar K and Mondal, Arnab Kumar and Ghosh, Arna and Richards, Blake},
  journal={Advances in Neural Information Processing Systems},
  volume={35},
  pages={17626--17638},
  year={2022}
}

@inproceedings{tsitsulin2023unsupervised,
    author    = "Tsitsulin, A. and Munkhoeva, M. and Perozzi, B.",
    title     = "Unsupervised embedding quality evaluation",
    booktitle = "Proceedings of the 40th International Conference on Machine Learning",
    pages     = "10929--10974",
    year      = "2023",
    publisher = "PMLR",
    url       = "https://arxiv.org/pdf/2305.16562",
}

@book{yukich2006probability,
  title={Probability theory of classical Euclidean optimization problems},
  author={Yukich, Joseph E},
  year={2006},
  publisher={Springer}
}

@online{dllllb2024gender,
    author    = "dllllb",
    title     = "Transactions Gender Prediction Dataset",
    year      = "2024",
    url       = "https://huggingface.co/datasets/dllllb/transactions-gender",
    note      = "Accessed via HuggingFace Datasets"
}

@online{dllllb2024agegroup,
    author    = "dllllb",
    title     = "Age Group Prediction from Transactions Dataset",
    year      = "2024",
    url       = "https://huggingface.co/datasets/dllllb/age-group-prediction",
    note      = "Accessed via HuggingFace Datasets"
}

@misc{ghrist_barcodes_2008,
  author        = {Ghrist, Robert},
  title         = {Barcodes: The Persistent Topology of Data},
  howpublished  = {\url{https://doi.org/10.1090/S0273-0979-07-01191-3}},
  year          = {2008},
  note          = {Accessed: 2025-06-04}
}

@misc{carlsson_topology_2009,
  author        = {Carlsson, Gunnar},
  title         = {Topology and Data},
  howpublished  = {\url{https://doi.org/10.1090/S0273-0979-09-01249-X}},
  year          = {2009},
  note          = {Accessed: 2025-06-04}
}

@inproceedings{babaev2022coles,
    author    = "Babaev, D. and Ovsov, N. and Kireev, I. and Gusev, G.",
    title     = "CoLES: Contrastive learning for event sequences with self-supervision",
    booktitle = "Proceedings of the 2022 International Conference on Management of Data",
    pages     = "1190--1199",
    year      = "2022",
    publisher = "ACM",
    url       = "https://arxiv.org/pdf/2002.08232",
}

@article{rendle2012bpr,
  title={BPR: Bayesian personalized ranking from implicit feedback},
  author={Rendle, Steffen and Freudenthaler, Christoph and Gantner, Zeno and Schmidt-Thieme, Lars},
  journal={arXiv preprint arXiv:1205.2618},
  year={2012}
}

@article{harper2015movielens,
  title={The movielens datasets: History and context},
  author={Harper, F Maxwell and Konstan, Joseph A},
  journal={Acm transactions on interactive intelligent systems (tiis)},
  volume={5},
  number={4},
  pages={1--19},
  year={2015},
  publisher={Acm New York, NY, USA}
}

@incollection{klenitskiy2025encode,
  title={Encode Me If You Can: Learning Universal User Representations via Event Sequence Autoencoding},
  author={Klenitskiy, Anton and Fatkulin, Artem and Denisova, Daria and Pembek, Anton and Vasilev, Alexey},
  booktitle={Proceedings of the Recommender Systems Challenge 2025},
  pages={26--30},
  year={2025}
}

@incollection{dkabrowski2025synerise,
  title={The SYNERISE dataset: An E-Commerce Dataset for Sequential Recommendation, Universal Behavior Modeling and Deep Relational Learning},
  author={D{\k{a}}browski, Jacek and Janicka, Maria and Sienkiewicz, {\L}ukasz and Stomfai, Gergely and Dietmar, Jannach and Barile, Francesco and Polignano, Marco and Pomo, Claudio and Srivastava, Abhishek},
  booktitle={Proceedings of the Recommender Systems Challenge 2025},
  pages={1--6},
  year={2025}
}

@inproceedings{dabrowski2025recsys,
  title={RecSys Challenge 2025: Universal Behavioral Profiles for Recommender Systems},
  author={Dabrowski, Jacek and Janicka, Maria and Sienkiewicz, Lukasz and Stomfai, Gergely and Jannach, Dietmar and Barile, Francesco and Polignano, Marco and Pomo, Claudio and Srivastava, Abhishek},
  booktitle={Proceedings of the Nineteenth ACM Conference on Recommender Systems},
  pages={1389--1393},
  year={2025}
}

@inproceedings{hadsell2006dimensionality,
  title={Dimensionality Reduction by Learning an Invariant Mapping},
  author={Hadsell, Raia and Chopra, Sumit and LeCun, Yann},
  booktitle={IEEE Computer Society Conference on Computer Vision and Pattern Recognition (CVPR)},
  year={2006}
}
